\documentclass{llncs}
\usepackage{graphicx}
\usepackage{amsmath,amssymb} 
\usepackage{subfigure}
\usepackage{algorithm}
\usepackage{algorithmic}
\usepackage{multirow}
\usepackage{color}
\usepackage[width=122mm,left=12mm,paperwidth=146mm,height=193mm,top=12mm,paperheight=217mm]{geometry}

\DeclareMathOperator*{\argmin}{argmin}
\begin{document}
\def\ECCV16SubNumber{1560}  

\title{MARLow: A Joint Multiplanar Autoregressive and Low-Rank Approach for Image Completion} 



\author{Mading Li${}^{1}$, Jiaying Liu${}^{1}$, Zhiwei Xiong${}^{2}$, Xiaoyan Sun${}^{2}$, Zongming Guo${}^{1}$}
\institute{${}^{1}$Institute of Computer Science and Technology, Peking University\\
${}^{2}$Microsoft Research Asia\\}

\maketitle

\begin{abstract}
In this paper, we propose a novel multiplanar autoregressive (AR) model to exploit the correlation in cross-dimensional planes of a similar patch group collected in an image, which has long been neglected by previous AR models. On that basis, we then present a joint multiplanar AR and low-rank based approach (\textbf{MARLow}) for image completion from random sampling, which exploits the nonlocal self-similarity within natural images more effectively. Specifically, the multiplanar AR model constraints the local stationarity in different cross-sections of the patch group, while the low-rank minimization captures the intrinsic coherence of nonlocal patches. The proposed approach can be readily extended to multichannel images (e.g. color images), by simultaneously considering the correlation in different channels. Experimental results demonstrate that the proposed approach significantly outperforms state-of-the-art methods, even if the pixel missing rate is as high as 90\%.
\end{abstract}

\section{Introduction}
Image restoration aims to recover original images from their low-quality observations, whose degradations are mostly generated by defects of capturing devices or error prone channels. It is one of the most important techniques in image/video processing, and low-level computer vision. In our work, we mainly focus on an interesting problem: image completion from random sampling, which has attracted many researchers' attention \cite{chierchia2014a,he2014iterative,heide2015fast,ji2010robust,liu2009tensor,zhang2012matrix,zhang2014image,zhang2014novel}. The problem is to recover the original image from its degraded observation, which has missing pixels randomly distributed. Such problem is a typical ill-posed problem, and different kinds of image priors have been employed.

One of the most commonly used image priors is the nonlocal prior \cite{buades2005review}, also known as the self-similarity property of natural images.
\begin{figure}[htbp]
  \centering
    \includegraphics[width=82mm]{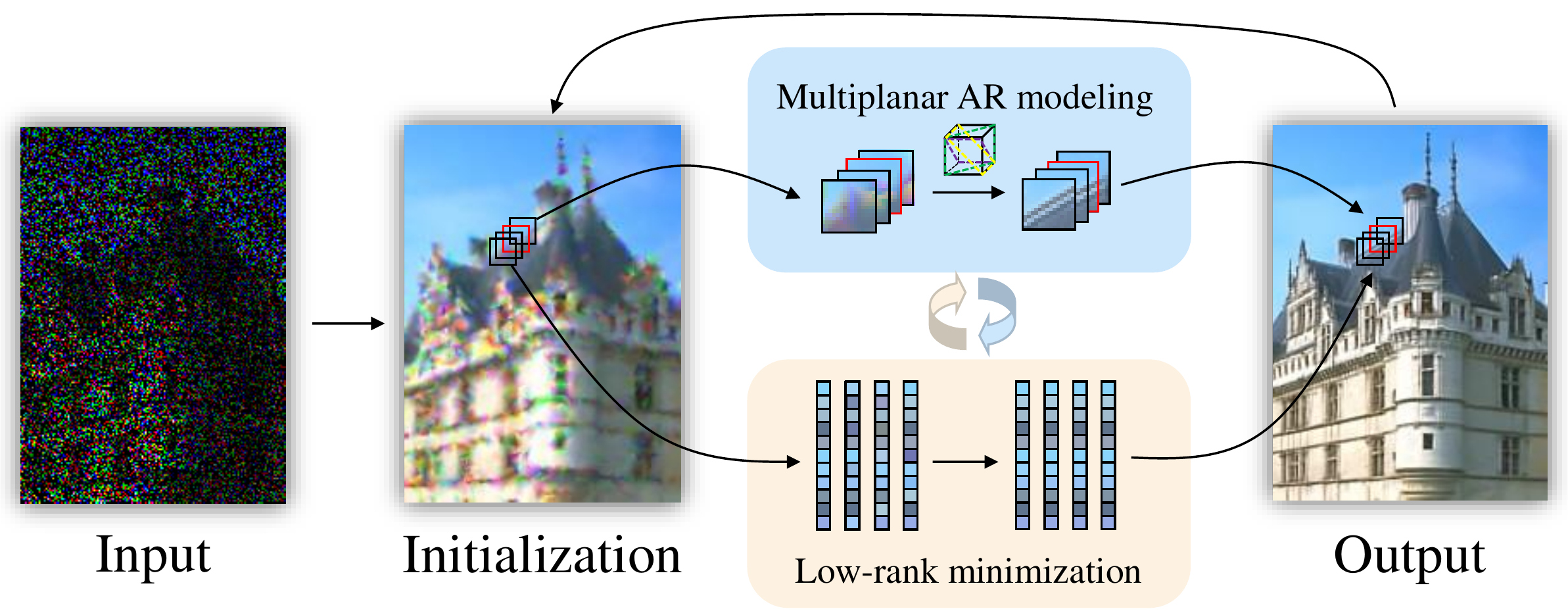}
    \vspace{-4mm}
  \caption{The framework of the proposed image completion method MARLow. After obtaining an initialization of the input image, similar patches are collected. Then, the joint multiplanar autoregressive and low-rank approach is applied on grouped patches. After all patches are processed, overlapped patches are aggregated into a new intermediate image, which can be used as the input for the next iteration.}
    \vspace{-7mm}
  \label{teaser}
\end{figure}
Such prior reflects the fact that there are many similar contents frequently repeated in the whole image, which can be well utilized in image completion. A classic way is to process the collected similar patch groups. The reason is that similar degraded patches contain complementary information for each other, which contributes to the completion. According to the manipulation scheme applied to the patch group, there are generally two kinds of methods in the literature:

\emph{\textbf{Cube-based methods}} stack similar patches directly, and then manipulate the data cube. The well-known denoising method Block-Matching and 3D filtering (BM3D) algorithm \cite{dabov2007image} is one of the most representative cube-based methods, which performs a 1D transform on each dimension of the data cube. The idea has been widely studied, and many extensions have been presented \cite{dabov2007video,maggioni2012video,zhang2014image}. These methods perform a global optimization on the data cube, neglecting the local structures inside the cube. Also, they process the data cube along each dimension, failing to consider the correlation that exists in cross-dimensional planes of the data cube. In this paper, we propose a multiplanar autoregressive (AR) model to address these problems. Specifically, the multiplanar AR model is to constrain the local stationarity in different sections of the data cube. Nonetheless, the multiplanar AR model is not good at smoothing the intrinsic structure of similar patches.

\emph{\textbf{Matrix-based methods}} stretch similar patches into vectors, which are spliced to form a data matrix. Two popular approaches, sparse coding and low-rank minimization, can be applied to such matrices. For sparse coding, the sparse coefficients of each vector in the matrix should be similar. This amounts to restricting the number of nonzero rows of the sparse coefficient matrix \cite{mairal2009nonlocal,zhou2012nonparametric}. Zhang \emph{et al.} \cite{zhang2014group} presented a group-based sparse representation method, which regards similar patch groups as its basic units. For low-rank minimization, since the data matrix is constructed by similar vectors, the rank of its underlying clean matrix to be recovered should be low. By minimizing the rank of the matrix, inessential contents (\emph{e.g.} the noise) of the matrix can be eliminated \cite{dong2013nonlocal,ji2010robust}. However, in image completion from random samples, such methods may excessively smooth the result, since they only consider the correlation of pixels at the same location of different patches. Also, unlike stacking similar patches directly, representing image patches by vectors shatters the local information stored in image patches.

Upon these analyses, these two kinds of methods seem to be relatively complemented. Thus, motivated by combining the merits of cube-based and matrix-based methods, we present a joint multiplanar autoregressive and low-rank approach (\textbf{MARLow}) for image completion (Figure \ref{teaser}). Instead of performing a global optimization on the data cube grouped by similar patches, we propose the concept of the multiplanar autoregressive model to exploit the local stationarity on different cross-sections of the data cube. Meanwhile, we jointly consider the matrix grouped by stretched similar patches, in which the intrinsic content of similar patches can be well recovered by low-rank minimization. In summary, our contributions lie in three aspects:

    \vspace{-3mm}
\begin{itemize}
\item We propose the concept of multiplanar autoregressive model, to characterize the local stationarity of cross-dimensional planes in the patch group.
\item We present a joint multiplanar autoregressive and low-rank approach (MARLow) for image completion from random sampling, along with an efficient alternating optimization method.
\item We extend our method to multichannel images by simultaneously considering the correlation in different channels, presenting encouraging performance.
\end{itemize}


\section{Related Work}
\label{related work}
In this section, we briefly review and discuss the existing literature that closely relates to the proposed method, including approaches associated with the autoregressive model and low-rank minimization.

\noindent
\textbf{Autoregressive model} ~~ The autoregressive (AR) model has been extensively studied in the last decades. AR model refers to modeling a pixel as the linear combination of its supporting pixels, usually its known neighboring pixels. Based on the assumption that natural images have the property of local stationarity, pixels in a local area share the same AR parameters, \emph{i.e.} the weight for each neighbor. AR parameters are often estimated from the low-resolution image \cite{li2001new,zhang2008image}. Dong \emph{et al.} proposed a nonlocal AR model \cite{dong2013sparse} using nonlocal pixels as supporting pixels, which is taken as a data fidelity constraint. The 3DAR model has been proposed to detect and interpolate the missing data in video sequences \cite{goh1996bidirectional,kokaram1994detection}. Since video sequences have the property of temporal smoothness, AR model can be extended to temporal space by combining the local statistics in the single frame. Different from approaches mentioned above, we focus on different cross-sections of the data cube grouped by similar nonlocal patches in a single image and constrain the local stationarity inside different planes in the data cube simultaneously (Figure \ref{3DARmodel}).

\noindent
\textbf{Low-rank minimization} ~~ As a commonly used tool in image completion, low-rank minimization aims to minimize the rank of an input corrupted matrix. It can be used for recovering/completing the intrinsic content of a degraded potentially low-rank matrix. The original low-rank minimization problem is NP-hard, and cannot be solved efficiently. Cand\`{e}s and Recht \cite{candes2009exact} proposed to relax the problem by using nuclear norm of the matrix, \emph{i.e.} the sum of singular values, which has been widely used in low-rank minimization problems since then. As proposed in \cite{dong2013nonlocal,ji2010robust}, similar patches in images/videos are collected to form a potentially low-rank matrix. Then, the nuclear norm of the matrix is minimized. Zhang \emph{et al.} further presented the truncated nuclear norm \cite{zhang2012matrix}, minimizing the sum of small singular values. Ono \emph{et al.} \cite{ono2014cartoon} proposed the block nuclear norm, leading to a suitable characterization of the texture component. Low-rank minimization can also be used on tensor completion. Liu \emph{et al.} \cite{liu2009tensor} regarded the whole input degraded color image as a potentially low-rank tensor, and defined the trace norm of tensors by extending the nuclear norm of matrices. However, most general natural images are not potentially low-rank. Thus, Chen \emph{et al.} \cite{chen2014simultaneous} attempted to recover the tensor while simultaneously capturing the underlying structure of it. In our work, we apply the nuclear norm of matrices, and we use singular value thresholding (SVT) method \cite{cai2010singular} to solve the low-rank minimization problem. Jointly combined with our multiplanar model (as elaborated in Section \ref{sec3}), our method produces encouraging image completion results.

\section{The Proposed Image Completion Method}
\label{sec3}
As discussed in previous sections, cube-based methods and matrix-based methods have their drawbacks, and they complement each other in some sense. In this section, we introduce the proposed multiplanar AR model to utilize information from cross-sections of the data cube grouped by similar patches. Moreover, combined with low-rank minimization, we present the joint multiplanar autoregressive and low-rank approach (MARLow) for image completion. At the end of this section, we extend the proposed method to multichannel images. For an input degraded image, we first conduct a simple interpolation-based initialization on it (see Figure \ref{teaser}), to provide enough information for patch grouping.

\subsection{Multiplanar AR Model}
\begin{figure}[htbp]
  \centering
  \subfigure[\scriptsize{Cross-sections of a data cube}]{
    \includegraphics[width=0.37\linewidth]{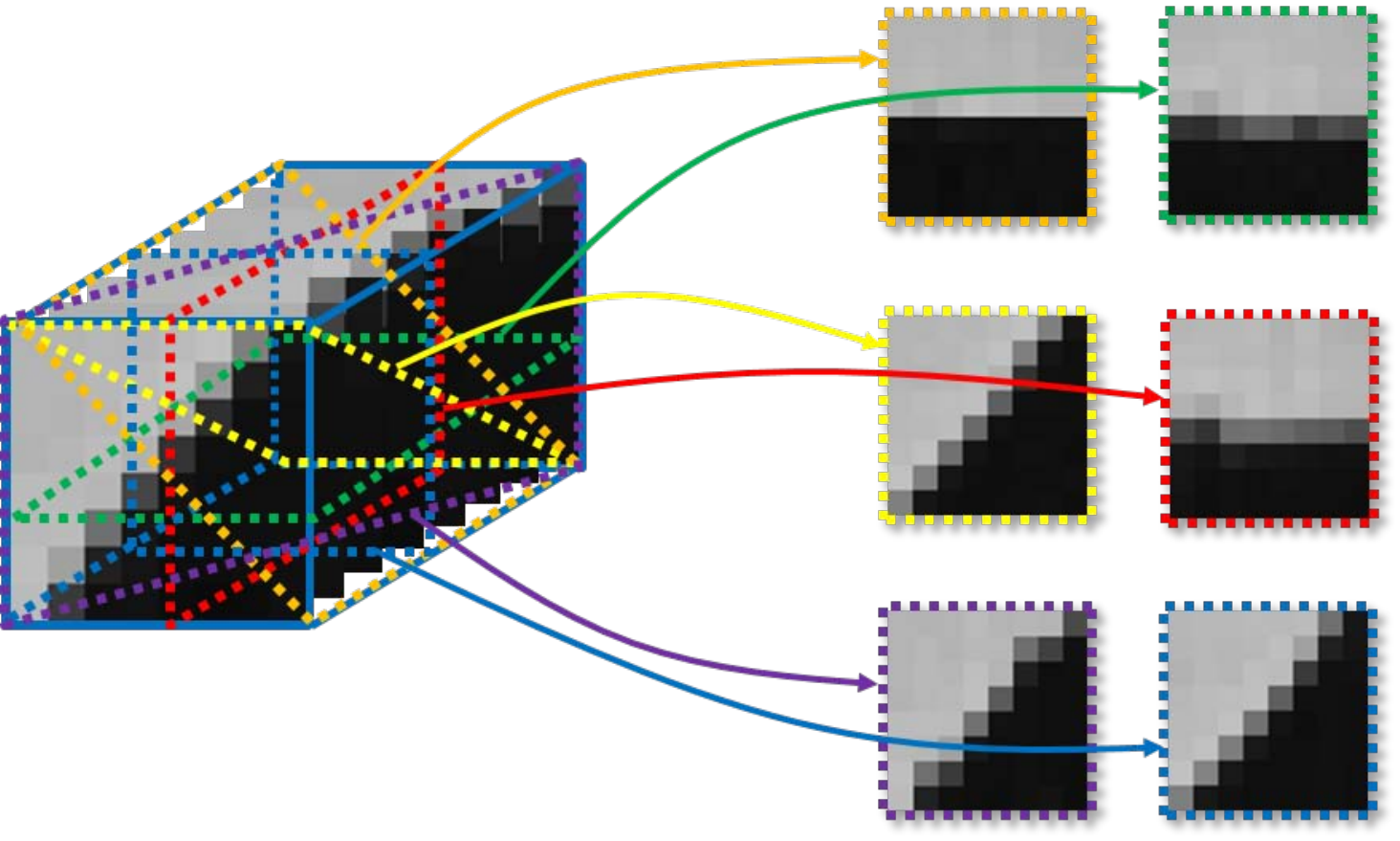}}
  \subfigure[\scriptsize{A multiplanar AR model}]{
    \includegraphics[width=0.29\linewidth]{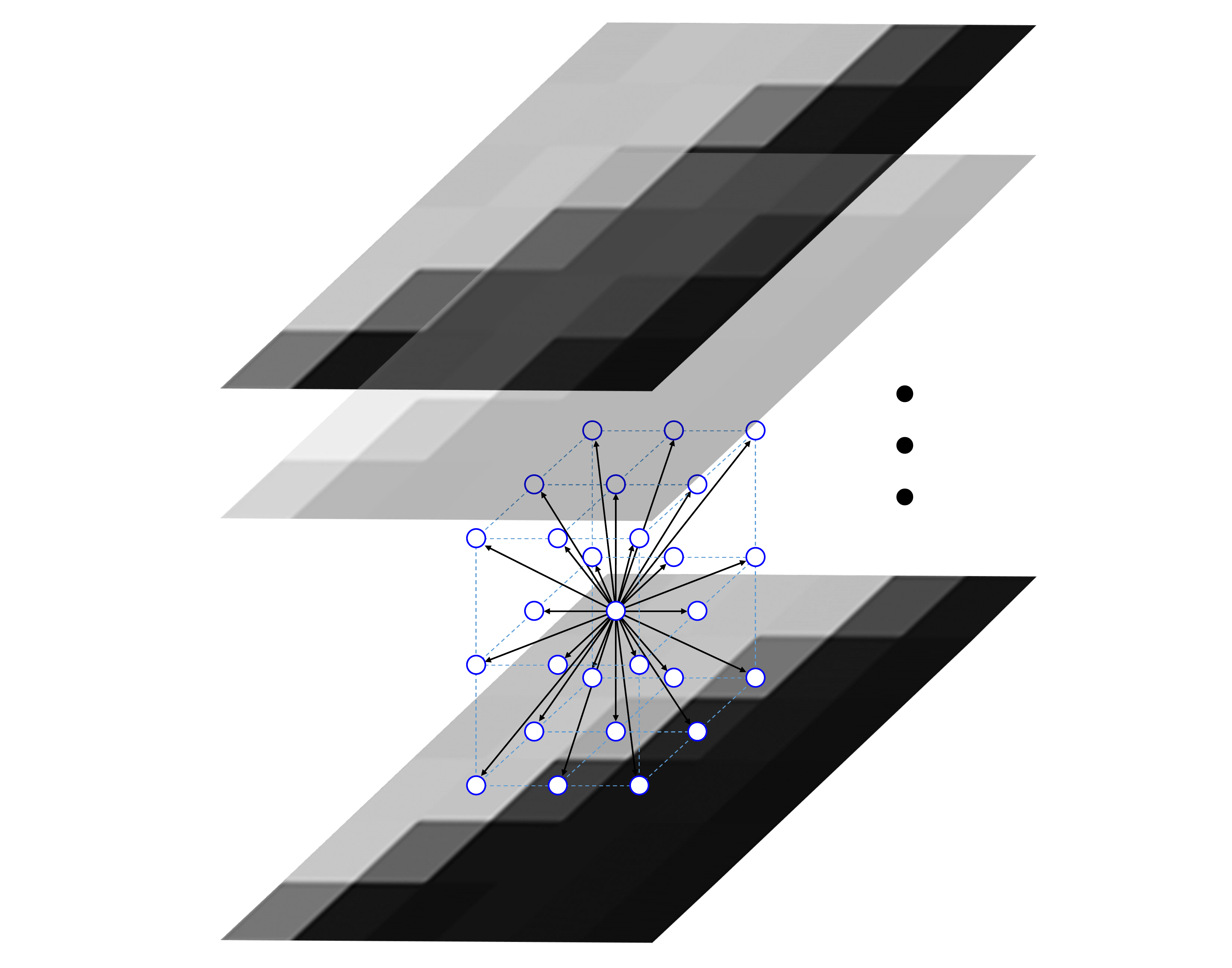}}
    \vspace{-4mm}
  \caption{(a) Different cross-sections of a data cube grouped by similar patches also possess local stationarity, which can be well processed by AR models. (b) White dots represent pixels (\emph{i.e.} small rectangles in $6\times 6$ image patches). Black arrows connect the center pixel of a multiplanar AR model with its supporting pixels.}
    \vspace{-6mm}
  \label{3DARmodel}
\end{figure}

Considering a reference patch of size $n\times n$, we collect its similar nonlocal patches. For a data cube grouped by similar patches, we observe its different cross-sections (cross-dimensional planes). As shown in Figure \ref{3DARmodel} (a), different cross-sections of the data cube possess local stationarity. Since AR models can measure the local stationarity of image signals, we naturally extend the conventional AR model to the multiplanar AR model to measure cross-dimensional planes.

Generally, the conventional AR model is defined as
\begin{equation}
X(i,j)=\sum_{(m,n)\in\mathcal{N}}X(i+m,j+n) \cdot \varphi(m,n)+\sigma(i,j),
\end{equation}
where $X(i,j)$ represents the pixel located at $(i,j)$. $X(i+m,j+n)$ is the supporting pixel with spatial offset $(m,n)$, while $\varphi(m,n)$ is the corresponding AR parameter. $\mathcal{N}$ is the set of supporting pixels' offsets and $\sigma(i,j)$ is the noise.

In our work, the multiplanar AR model consists supporting pixels from different cross-dimensional planes (as illustrated in Figure \ref{3DARmodel} (b)). For a data cube grouped by similar patches of an image patch located at $i$, the multiplanar AR model of pixel $X_i(j,k,l)$ with offset $(j,k,l)$ in the data cube is defined as
\begin{equation}
\label{3DAR}
X_i(j,k,l) = \sum_{m \in {\mathcal{N}_1}} \sum_{\tiny\begin{array}{c}(p,q)\\ \in {\mathcal{N}_2}\end{array}} Y_i(j + m,k + p,l + q) \cdot \varphi_i (m,p,q) + \sigma_i(j,k,l),
\end{equation}
where $\mathcal{N}_1$ represents the set of supporting pixels' planar offsets and $\mathcal{N}_2$ represents the set of supporting pixels' spatial offsets (assuming the order of the multiplanar AR model $N_{order} = |\mathcal{N}_1|\times|\mathcal{N}_2|$). $Y_i(j + m,k + p,l + q)$ is the supporting pixel with offset $(m,p,q)$ in the data cube and $\varphi_i (m,p,q)$ is the corresponding AR parameter. $\sigma_i(j,k,l)$ is the noise. $Y$ is the initialization of the input image $X$. The reason we use $Y$ here is that it is difficult to find enough known pixels to support the multiplanar AR model under high pixel missing rate.

For an $n\times n$ patch, assuming $N$ patches are collected, the aforementioned multiplanar AR model can be transformed into a matrix form, that is,
\begin{equation}
X_i = T_i(Y)\cdot\varphi_i+\sigma_i,
\end{equation}
where $X_i\in\mathbb{R}^{(n^2\times N)\times1}$ is a vector containing all modeled pixels. $T_i(\cdot)$ represents the operation that extract supporting pixels for $X_i$. Each row of $T_i(Y)\in\mathbb{R}^{(n^2\times N)\times N_{order}}$ contains values of supporting pixels of each pixel and $\varphi_i\in\mathbb{R}^{N_{order}\times 1}$ is the multiplanar AR parameter vector.

Thus, the optimization problem for $X_i$ and $\varphi_i$ can be formulated as follows,
\begin{equation}
\mathop {\argmin }_{X_i,\varphi_i } {\left\| {{X_i} - {T_i}(Y)\cdot{\varphi _i}} \right\|_F^2}.
\end{equation}
where $\|\cdot\|_F$ is the Frobenius norm. In order to enhance the stability of the solution, we introduce the Tikhonov regularization to solve this problem. Specifically, a regularization term is included in the minimization problem, forming the following regularized least-square problem
\begin{equation}
\label{multiplanarAR}
\mathop {\argmin }_{X_i,\varphi_i } {\left\| {{X_i} - {T_i}(Y)\cdot{\varphi _i}} \right\|_F^2+ \left\| {\Gamma \cdot{\varphi _i}} \right\|_F^2},
\end{equation}
where $\Gamma = \alpha I$ and $I$ is an identity matrix.

\subsection{MARLow}
Since the multiplanar AR model is designed to constrain a pixel with its supporting pixels on different cross-sections of the patch group, it can deal more efficiently with local image structures. For instance, assume there is an edge on an image that is severely degraded, with only a few pixels on it. After collecting similar patches, low-rank minimization or other matrix-based methods may regard the remaining pixels as noises and remove them. However, with the multiplanar AR model, these pixels can be used to constrain each other and strengthen the underlying edge. Nevertheless, AR models are not suitable for smoothing the intrinsic structure, while low-rank minimization methods specialize in it. So we propose to combine the multiplanar AR model with low-rank minimization (MARLow) as follows,
\begin{equation}
\begin{split}
\label{objective}
\mathop {\argmin }_{X_i,\varphi_i } & \left\| {{X_i} - {T_i}{(Y)}\cdot{\varphi _i}} \right\|_F^2 + \left\| {\Gamma\cdot {\varphi _i}} \right\|_F^2\\
&+ \mu\left(\left\| {{R_i}(X) - {R_i}(Y)} \right\|_F^2 + {{\left\| {{R_i}(X)} \right\|}_*}\right),
\end{split}
\end{equation}
where the last part is the low-rank minimization term restricting the fidelity while minimizing the nuclear norm (\emph{i.e.} $\|\cdot\|_*$) of the data matrix. $R_i(\cdot)$ is an extraction operation that extracts similar patches of the patch located at $i$. $R_i(X) = [X_{i_1},X_{i_2},...,X_{i_N}]\in\mathbb{R}^{n^2\times N}$ is similar patch group of the reference patch $X_{i_1}$, and $R_i(Y) = [Y_{i_1},Y_{i_2},...,Y_{i_N}]\in\mathbb{R}^{n^2\times N}$ represents the corresponding patch group extracted from $Y$.

\begin{figure}[hbtp]
\centering
  \vspace{-6mm}
  \subfigure[Low-rank]{
    \includegraphics[width=30mm]{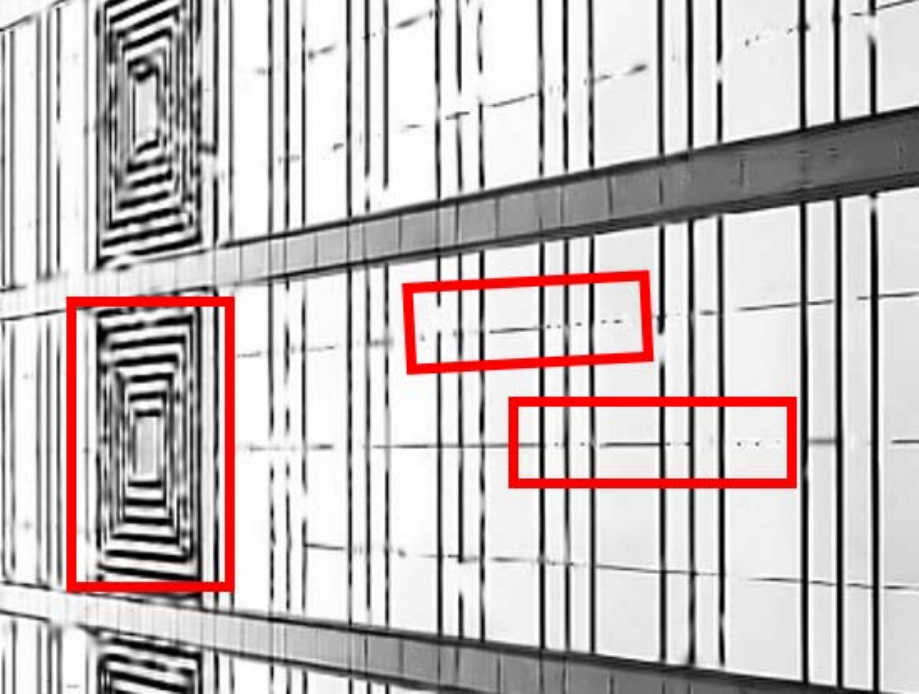}}\hspace{1.5mm}
  \subfigure[MARLow]{
    \includegraphics[width=30mm]{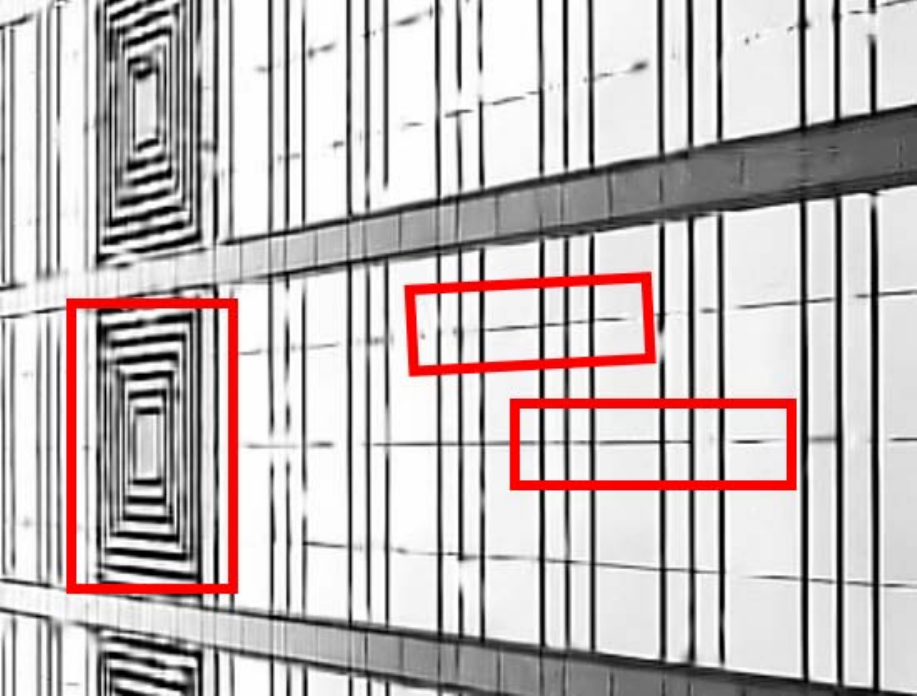}}
  \vspace{-4mm}
  \caption{(a) Completion result by low-rank without our multiplanar AR model (19.48 dB/0.8933); (b) Completion result generated by MARLow (\textbf{19.95 dB}/\textbf{0.8980}). Clearly, the result obtained by MARLow has higher visual quality, which can be observed in marked regions.}
  \vspace{-8mm}
  \label{withandwithout3DAR}
\end{figure}

Figure \ref{withandwithout3DAR} presents the completion results by using only low-rank without the multiplanar AR model, and by MARLow. From the figure, we can see that MARLow can effectively connect fractured edges.

\subsection{Multichannel Image Completion}
For multichannel images, instead of applying the straightforward idea, that is, the separate procedure (\emph{i.e.} processing different channels separately and combining the results afterward), we present an alternative scheme to simultaneously process different channels. At first, we collect similar patches of size $n\times n\times h$ (where $h$ represents the number of channels) in a multichannel image. After that, each patch group is processed by simultaneously considers all channels. Specifically, the collected patches can be formed into $h$ data cubes by stacking slices (of size $n\times n\times 1$) in the corresponding channel of different patches. For multiplanar AR model, the minimization problem in Equation (\ref{multiplanarAR}) turns into
\begin{equation}
\mathop {\argmin }_{X_i^k,\varphi_i^k} \sum_{1\leq k\leq h}\left({\left\| {{X_i^k} - {T_i^k}(Y)\cdot{\varphi _i^k}} \right\|_F^2+ \left\| {\Gamma\cdot {\varphi _i^k}} \right\|_F^2}\right)
\end{equation}

For low-rank minimization, $N$ collected patches are formed into a potentially low-rank data matrix of size $(n^2\times h)\times N$ by representing each patch as a vector.

Taking an RGB image for an example, in patch grouping, we search for similar patches using reference patches with the size $n\times n \times 3$. The multichannel image completion problem can be solved by minimizing
\begin{equation}
\begin{split}
\label{objectiveColor}
&\mathop {\argmin }_{X^C_i,\varphi_i^C } \left\| {{X_i^C} - {T_i^C}{(Y^C)}\cdot{\varphi^C_i}} \right\|_F^2 + \left\| {\Gamma\cdot {\varphi^C_i}} \right\|_F^2 \quad\quad\\
& \ \ \ + \mu\left(\left\| {{R_i^C}(X^C) - {R_i^C}(Y^C)} \right\|_F^2 + {{\left\| {{R_i^C}(X^C)} \right\|}_*}\right),
\end{split}
\end{equation}
where
\begin{equation*}
X_i^C=
\begin{bmatrix}
X_i^R \\
X_i^G \\
X_i^B \\
\end{bmatrix}\in\mathbb{R}^{(n^2\times N\times 3)\times1},
\end{equation*}
\begin{equation*}
\varphi_i^C=
\begin{bmatrix}
\varphi_i^R \\
\varphi_i^G \\
\varphi_i^B \\
\end{bmatrix}\in\mathbb{R}^{(N_{order}\times3)\times1},
\end{equation*}
\begin{equation*}
T_i^C(Y^C) =
\begin{bmatrix}
T_i(Y^R)\hspace{-3mm}&0&0\\
0&T_i(Y^G)\hspace{-4mm}&0\\
0&0&T_i(Y^B)
\end{bmatrix}\in\mathbb{R}^{(n^2\times N\times3)\times(N_{order}\times3)},
\end{equation*}
\begin{equation*}
R_i^C(X^C) =
\begin{bmatrix}
X_{i_1}^R&X_{i_2}^R&\cdots&X_{i_N}^R\\
X_{i_1}^G&X_{i_2}^G&\cdots&X_{i_N}^G\\
X_{i_1}^B&X_{i_2}^B&\cdots&X_{i_N}^B
\end{bmatrix}\in\mathbb{R}^{(n^2\times3)\times N},
\end{equation*}
\begin{equation*}
R_i^C(Y^C) =
\begin{bmatrix}
Y_{i_1}^R&Y_{i_2}^R&\cdots&Y_{i_N}^R\\
Y_{i_1}^G&Y_{i_2}^G&\cdots&Y_{i_N}^G\\
Y_{i_1}^B&Y_{i_2}^B&\cdots&Y_{i_N}^B
\end{bmatrix}\in\mathbb{R}^{(n^2\times3)\times N}.
\end{equation*}

\begin{figure}[hbtp]
\centering
  \subfigure[25.77 dB]{
    \includegraphics[width=0.15\linewidth]{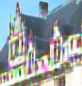}} 
  \subfigure[30.36 dB]{
    \includegraphics[width=0.15\linewidth]{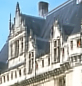}} 
  \subfigure[26.66 dB]{
    \includegraphics[width=0.17\linewidth]{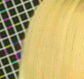}} 
  \subfigure[34.38 dB]{
    \includegraphics[width=0.17\linewidth]{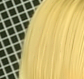}}
  \vspace{-4mm}
  \caption{(a) and (c) are obtained by the separate procedure. (b) and (d) are obtained by our multichannel image completion method.}
  \vspace{-6mm}
  \label{seperateandsimultaneous}
\end{figure}

The notations are given similarly as the preceding definitions. By utilizing the information in multichannel images, the patch grouping can be more precise. Furthermore, rich information in different channels can compensate for each other and constrain the completion result. Figure \ref{seperateandsimultaneous} illustrates the difference between processing different channels separately and simultaneously (with 80\% pixels missing). Compared with the separate procedure, the multichannel image completion approach can significantly improve the performance of our method. Thus, in Section \ref{experiment}, for those methods dedicated to gray-scale image completion, we do not apply the separate procedure to them to obtain color image completion results since it may be unfair. Instead, we compare our multichannel image completion method with other state-of-the-art color image completion methods.

\section{Optimization}
\label{optimization}
In this section, we present an alternating minimization algorithm to solve the minimization problems in Equation (\ref{objective}) and (\ref{objectiveColor}). Take Equation (\ref{objective}) for an example. We address each of the variable $X_i$ and $\varphi_i$ separately and present an efficient optimization algorithm.

When fixing $X_i$, the problem turns into
\begin{equation}
\label{updatephi}
\mathop {\argmin }_{\varphi_i } {\left\| {{X_i} - {T_i}{(Y)}\cdot{\varphi _i}} \right\|_F^2 + \left\| {\Gamma\cdot {\varphi _i}} \right\|_F^2},
\end{equation}
which is a standard regularized linear least square problem, and can be solved by ridge regression. The closed-form solution is given by
\begin{equation}
\label{calcphi}
\varphi_i = \left(\hat{Y}^T\hat{Y}+\Gamma^T\Gamma\right)^{-1}\hat{Y}X_i,
\end{equation}
where $\hat{Y} = {T_i}{(Y)}$.

With $\varphi_i$ fixed, the problem for updating $X_i$ becomes
\begin{equation}
\label{updateX}
\begin{split}
&\mathop {\argmin }_{X_i} \left\| {{X_i} - {T_i}{(Y)}\cdot{\varphi _i}} \right\|_F^2 \quad\quad\quad \\
&\quad\quad\quad + \mu\left(\left\| {{R_i}(X) - {R_i}(Y)} \right\|_F^2 + {{\left\| {{R_i}(X)} \right\|}_*}\right).
\end{split}
\end{equation}

Here we notice that $X_i$ and $R_i(X)$ contain the same elements. Their only difference is the formation: $X_i$ is a vector and $R_i(X)$ is a matrix. Since we use Frobenius norm here, the value of the norm does not change if we reform the vector into a matrix form. So we reform $X_i$ into a matrix $M_i$ corresponding to $R_i(X)$ (in this way, $R_i(X)$ does not need to be reformed, and it can be represented by $M_i$ directly). The vector $T_i(Y)\cdot\varphi_i$ is also reformed into a matrix form, represented by $Y_{1_i}$. By denoting $Y_{2_i} = R_i(Y)$, we can get the simplified version of Equation (\ref{updateX}):
\begin{equation}
\mathop {\argmin }_{M_i} {\left\| {{M_i} - Y_{1_i}} \right\|_F^2 + \mu\left(\left\| {{M_i} - Y_{2_i}} \right\|_F^2 + {{\left\| M_i \right\|}_*}\right)}.
\end{equation}

It is a modified low-rank minimization problem and can be transformed into the following formation
\begin{equation}
\label{typicalLR}
\mathop {\argmin }_{M_i} {\left\| {{M_i} - Y_{i}'} \right\|_F^2 + \lambda{{\left\| M_i \right\|}_*}},
\end{equation}
where $Y_i'=(1-\lambda)Y_{1_i}+\lambda Y_{2_i}$ and $\lambda = \mu/(\mu+1)$. The problem now turns into a standard low-rank minimization problem \cite{cai2010singular}. Its closed-form solution is given as
\begin{equation}
\label{iteration}
M_i=S_\tau(Y_i'),
\end{equation}
where $S_\tau(\cdot)$ represents the soft shrinkage process.

With the input random sampled image $Y$ and the mask matrix $M_{mask}$ indicating known pixels ($0$s for missing pixels and $1$s for known pixels), our alternating minimization algorithm for image completion from random sampling can be summarized in Algorithm \ref{algorithm}.

\begin{algorithm}
\caption{A Joint Multiplanar Autoregressive and Low-Rank Approach for Image Completion}
\label{algorithm}
\begin{algorithmic}
\STATE \textbf{Input:} $Y$ and $M_{mask}$.\\
\STATE $X^0=Bilinear(Y,M_{mask})$.\\
\FOR {$i = 1$ to $maxIter$}
    \STATE Patch grouping.
    \FOR {each image patch group $X_k$ in $X^{(i-1)}$}
        \STATE Estimate $\varphi_k$ according to Equation (\ref{updatephi}).
        \STATE Estimate $X_k$ according to Equation (\ref{updateX}).
    \ENDFOR
    \STATE Estimate $X^{(i)}$ by aggregating all overlapped patches.
    \STATE $X^{(i)}(M_{mask}) = Y(M_{mask})$.
    \STATE $Y = X^{(i)}$.
\ENDFOR
\STATE \textbf{Output:} The restored image $X^{(i)}$.\\
\end{algorithmic}
\end{algorithm}
\vspace{-6mm}

\section{Experimental Results}
\label{experiment}

Experimental results of compared methods are all generated by the original authors' codes, with the parameters manually optimized. Both objective and subjective comparisons are provided for a comprehensive evaluation of our work. Peak Signal to Noise Ratio (PSNR) and structural similarity (SSIM) index are used to evaluate the objective image quality. In our implementation, if not specially stated, the size of each image patch is set to $8\times 8$ ($5\times5\times3$ in color images) with four-pixel (one-pixel in color images) overlap. The number of similar patches is set to $N = 64$ for gray-scale image and $N = 75$ for color image. Other parameters in our algorithm are empirically set to $\alpha = \sqrt{10}$, $\mu = 10$. Please see the electronic version for better visualization of the subjective comparisons. More results can be found in the supplementary materials.


\begin{table*}[htbp]
\footnotesize
  \centering
  \caption{PSNR (dB) and SSIM results of gray-scale image completion from different methods under 80\% missing rate. The best result in each case is highlighted in bold.}
    \begin{tabular}{|c|c|c|c|c|c|}
    \hline
    \textbf{Image} & \multicolumn{1}{|c}{\textbf{ISDSB}} & \multicolumn{1}{|c}{\textbf{BNN}} & \multicolumn{1}{|c}{\textbf{BPFA}} & \multicolumn{1}{|c}{\textbf{JSM}} & \multicolumn{1}{|c|}{\textbf{Proposed}} \\
    \hline\hline
    \textit{House} & 25.61/0.8052 & 27.76/0.8381 & 30.19/0.8717 & 33.00/0.8944 & \textbf{34.70}/\textbf{0.9070} \\
    \hline
    \textit{Lena} & 27.31/0.8142 & 28.58/0.8416 & 30.95/0.8794 & 31.49/0.8836 & \textbf{32.84}/\textbf{0.9043} \\
    \hline
    \textit{Cameraman} & 21.72/0.7584 & 22.65/0.7867 & 24.04/0.8082 & 25.18/0.8439 & \textbf{25.49}/\textbf{0.8581} \\
    \hline
    \textit{Pepper} & 27.06/0.8205 & 27.87/0.8389 & 29.85/0.8529 & 31.75/0.8664 & \textbf{32.59}/\textbf{0.8781} \\
    \hline\hline
    \textbf{Average} & 25.42/0.7996 & 26.72/0.8263 & 28.76/0.8530 & 30.35/0.8721 & \textbf{31.41}/\textbf{0.8869} \\
    \hline

    \end{tabular}%
  \label{graytable}%
\vspace{-3mm}
\end{table*}%

\begin{table*}[htbp]
\footnotesize
  \centering
  \caption{PSNR (dB) and SSIM results of color image completion from different methods under 80\% and 90\% missing rates. The best result in each case is highlighted in bold.}
    \begin{tabular}{|c|c|c|c|c|c|c|}
    \hline
    \textbf{Image} & \textbf{Ratio} & \textbf{FoE} & \textbf{ST-NLTV} & \textbf{GSR} & \textbf{BPFA} & \textbf{Proposed} \\
    \hline\hline
        \multirow{2}[4]{*}{\textit{Castle}} & 80\%  & 25.71/0.8424 & 26.60/0.8414 & 25.66/0.8588 & 29.22/0.9099 & \textbf{30.36}/\textbf{0.9124} \\
          & 90\%  & 23.38/0.7655 & 23.39/0.7507 & 22.99/0.7771 & 25.09/0.8189 & \textbf{26.55}/\textbf{0.8509} \\
    \hline
    \multirow{2}[4]{*}{\textit{Woman}} & 80\%  & 19.83/0.7948 & 21.37/0.7992 & 31.71/0.9460 & 30.33/0.9363 & \textbf{34.38}/\textbf{0.9561} \\
          & 90\%  & 17.16/0.6520 & 16.79/0.5973 & 20.02/0.8016 & 24.20/0.8640 & \textbf{30.12}/\textbf{0.9179} \\
    \hline
    \multirow{2}[4]{*}{\textit{Soldier}} & 80\%  & 23.95/0.8434 & 25.10/0.8320 & 25.54/0.8963 & 28.92/0.9352 & \textbf{30.78}/\textbf{0.9473} \\
          & 90\%  & 21.03/0.7237 & 20.44/0.6911 & 21.82/0.7922 & 24.09/0.8368 & \textbf{26.34}/\textbf{0.8893} \\
    \hline\hline
    \multicolumn{2}{|c|}{\textbf{Average}} & 21.84/0.7703 & 22.28/0.7520 & 24.62/0.8453 & 26.98/0.8835 & \textbf{29.76}/\textbf{0.9123} \\

    \hline
    \end{tabular}%
  \label{colortable}%
\vspace{-5mm}
\end{table*}%

\subsection{Gray-Scale Image Completion}
For gray-scale images, we compare our method with state-of-the-art gray-scale image completion methods BPFA \cite{zhou2012nonparametric}, BNN \cite{ono2014cartoon}, ISDSB \cite{he2014iterative}, and JSM \cite{zhang2014image}.
Table \ref{graytable} shows PSNR/SSIM results of different methods on test images with 80\% pixels missing. From Table \ref{graytable}, the proposed method achieves the highest PSNR and SSIM in all cases, which fully demonstrates the effectiveness of our method. Specifically, the improvement on PSNR is 1.06 dB and that on SSIM is 0.0148 on average compared with the second best algorithm (\emph{i.e.} JSM).

\begin{figure*}[hbtp]
\vspace{-3mm}
  \centering
  \includegraphics[width=120mm]{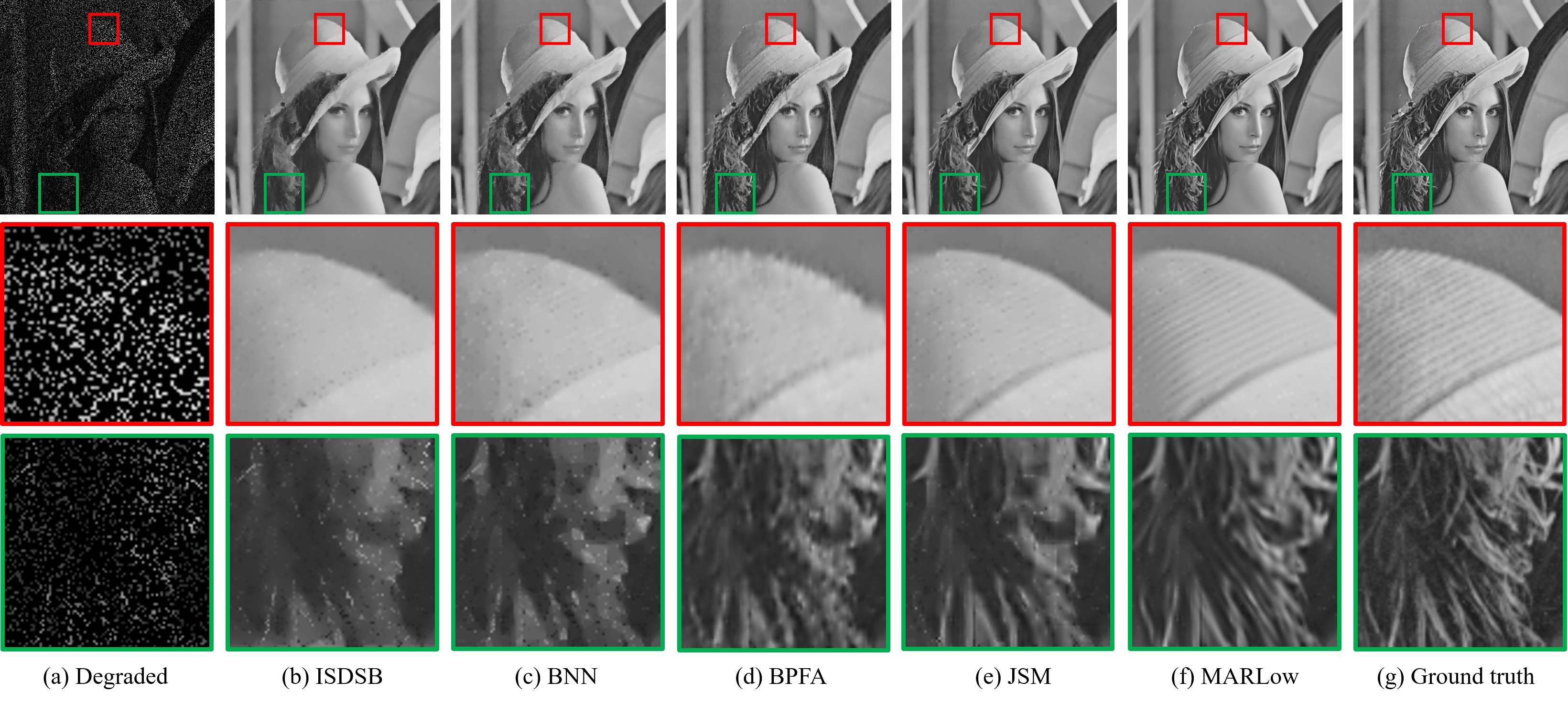}
\vspace{-4mm}
  \caption{Comparison of completion results of different methods with 80\% pixels missing. From left to right: the degraded image, results of ISDSB, BNN, BPFA, JSM, our method, and the ground truth. The second and third rows show the corresponding close-ups.}
\vspace{-6mm}
  \label{gray-80}
\end{figure*}

Figure \ref{gray-80} compares the visual quality of completion results for test images (with 80\% pixels missing). From Figure \ref{gray-80}, ISDSB and BNN successfully recover the boundaries of the image, but fail to restore rich details. BPFA performs better completion on image details. Nonetheless, there are plenty of noises along edges recovered by BPFA. At the first glance, the completion results of JSM and our method are both of high quality. However, if we get a closer look, it can be observed that there are isolated noises on image details (such as structures on Lena's hair and her hat) in the result generated by JSM. JSM also cannot recover tiny structures. Our method presents the best visual quality, especially on image details and edges.

\subsection{Color Image Completion}
We compare our method with state-of-the-art color image completion methods FoE \cite{roth2005fields}, BPFA \cite{zhou2012nonparametric}, GSR \cite{zhang2014group} and ST-NLTV \cite{chierchia2014a}.
Table \ref{colortable} lists PSNR/SSIM results of different methods on color images with 80\% and 90\% pixels missing. It is clear that the proposed method achieves the highest PSNR/SSIM in all cases. Compared with gray-scale images, our image completion method performs even better on color images judging from the average PSNR and SSIM. The proposed method outperforms the second best method (\emph{i.e.} BPFA) by 2.78 dB on PSNR and 0.0288 on SSIM. Note that, when tested on image \emph{Woman} with 90\% pixels missing, the PSNR and SSIM improvements achieved by our method over BPFA are 5.92 dB and 0.0539, respectively.

\begin{figure*}[hbtp]
\vspace{-3mm}
\centering
  \includegraphics[width=120mm]{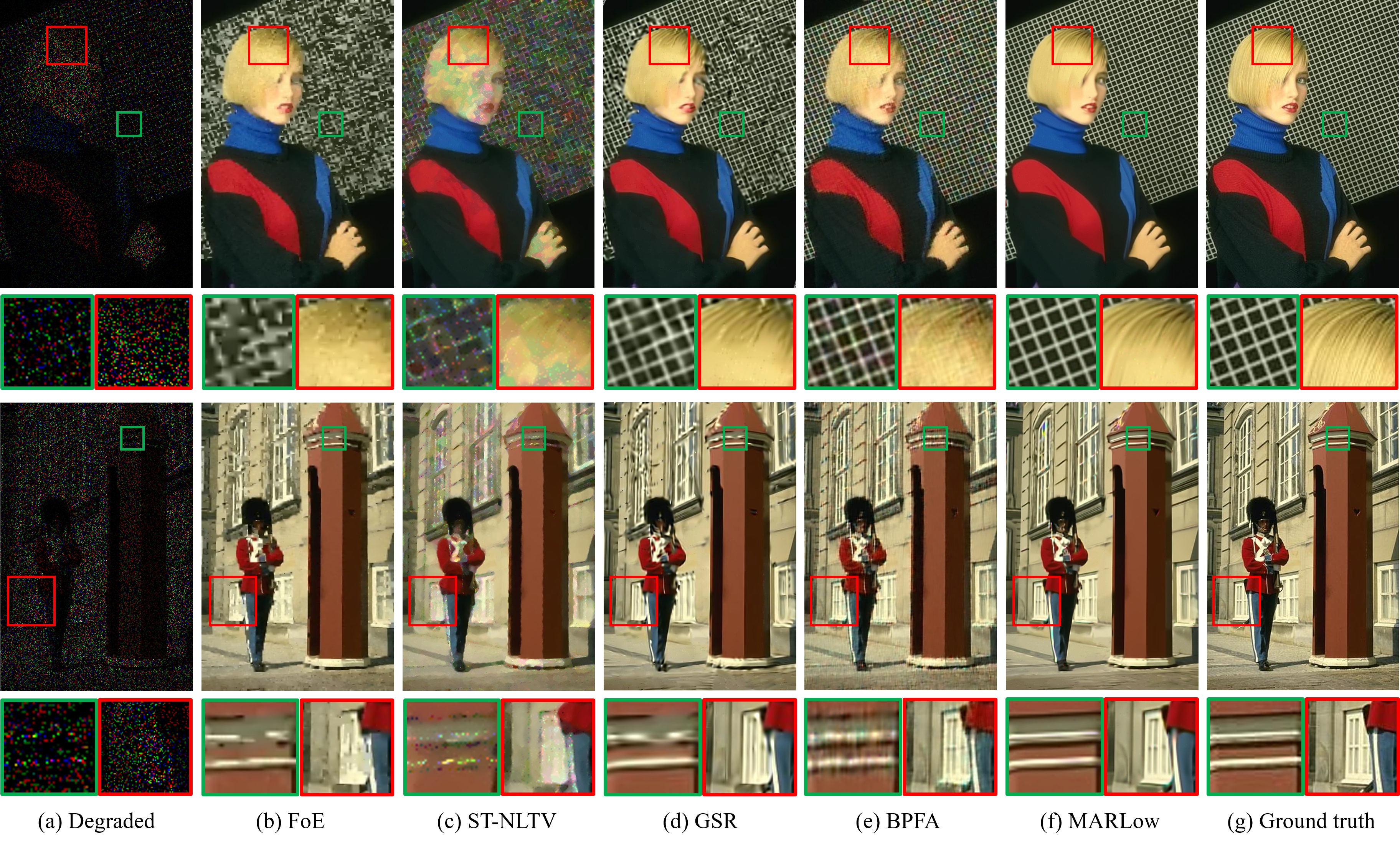}
\vspace{-4mm}
  \caption{Comparison of color image completion results of different methods with 90\% pixels missing. From left to right: the degraded image, results of FoE, ST-NLTV, GSR, BPFA and our method, the ground truth.}
\vspace{-6mm}
  \label{color-90}
\end{figure*}

Figure \ref{color-90} shows the visual quality of color image completion results for test images (with 90\% pixels missing). Apparently, all the comparing methods are doing great on flat regions. However, FoE and ST-NLTV cannot restore fine details. GSR is better on recovering details, but it generates noticeable artifacts around edges and fails to connect fractured edges. BPFA produces sharper edges, but its performance under higher missing rate is not satisfying. The result of our method is of the best visual quality, especially under higher missing rate.

\begin{figure*}[hbtp]
\vspace{-3mm}
  \centering
  \includegraphics[width=120mm]{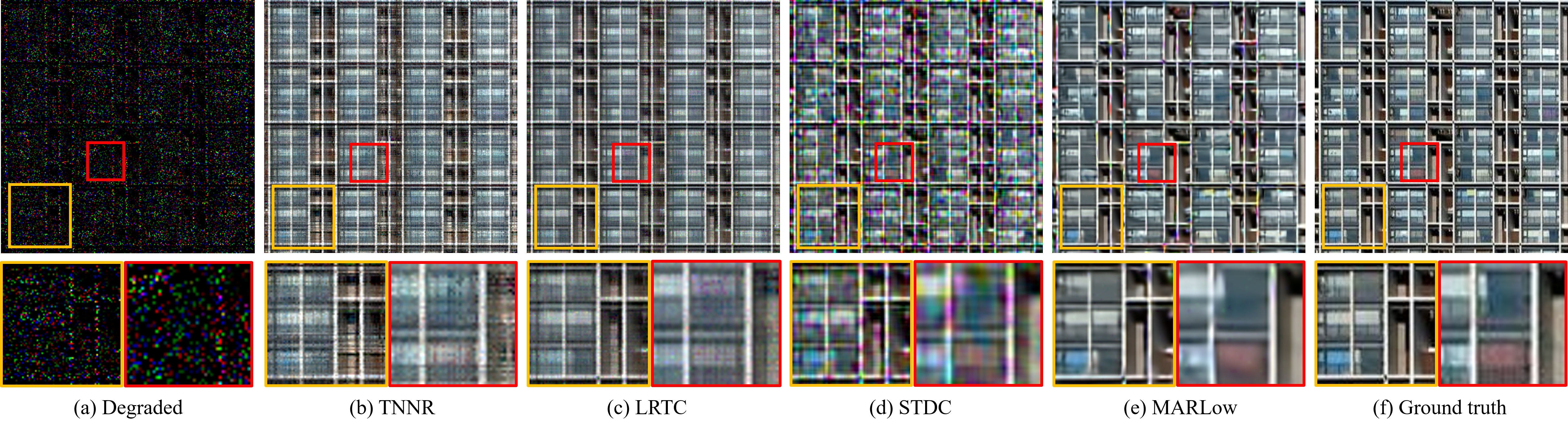}
\vspace{-4mm}
  \caption{Visual quality comparison of low-rank based methods. (a) The degraded image (with 90\% pixels missing, 6.12 dB/0.0292); (b) - (e) Completion results by TNNR (17.58 dB/0.7276), LRTC (19.78 dB/0.8226), STDC (17.65 dB/0.7839), and the proposed method (\textbf{21.71 dB}/\textbf{0.9021}). (f) The ground truth image.}
  \label{facade-color-80-90}
\vspace{-6mm}
\end{figure*}

We also compare our method with state-of-the-art low-rank matrix/tensor completion based methods TNNR \cite{zhang2012matrix}, LRTC \cite{liu2009tensor} and STDC \cite{chen2014simultaneous}. Since these methods regard the whole image as a potentially low-rank matrix, the input image should have strong correlations between its columns or rows. Thus, to be fair, we also test this kind of images to evaluate the performance of our method. From Figure \ref{facade-color-80-90}, TNNR and LRTC tends to erase tiny objects of the image, such as the colorful items (see the close-ups in Figure \ref{facade-color-80-90}). STDC imports noticeable noises into the whole image. The proposed method presents not only accurate completion on sharp edges, but also high-quality textures, exhibiting the best visual quality.

\subsection{Text Removal}

\begin{figure*}[hbtp]
\vspace{-3mm}
  \centering
  \includegraphics[width=120mm]{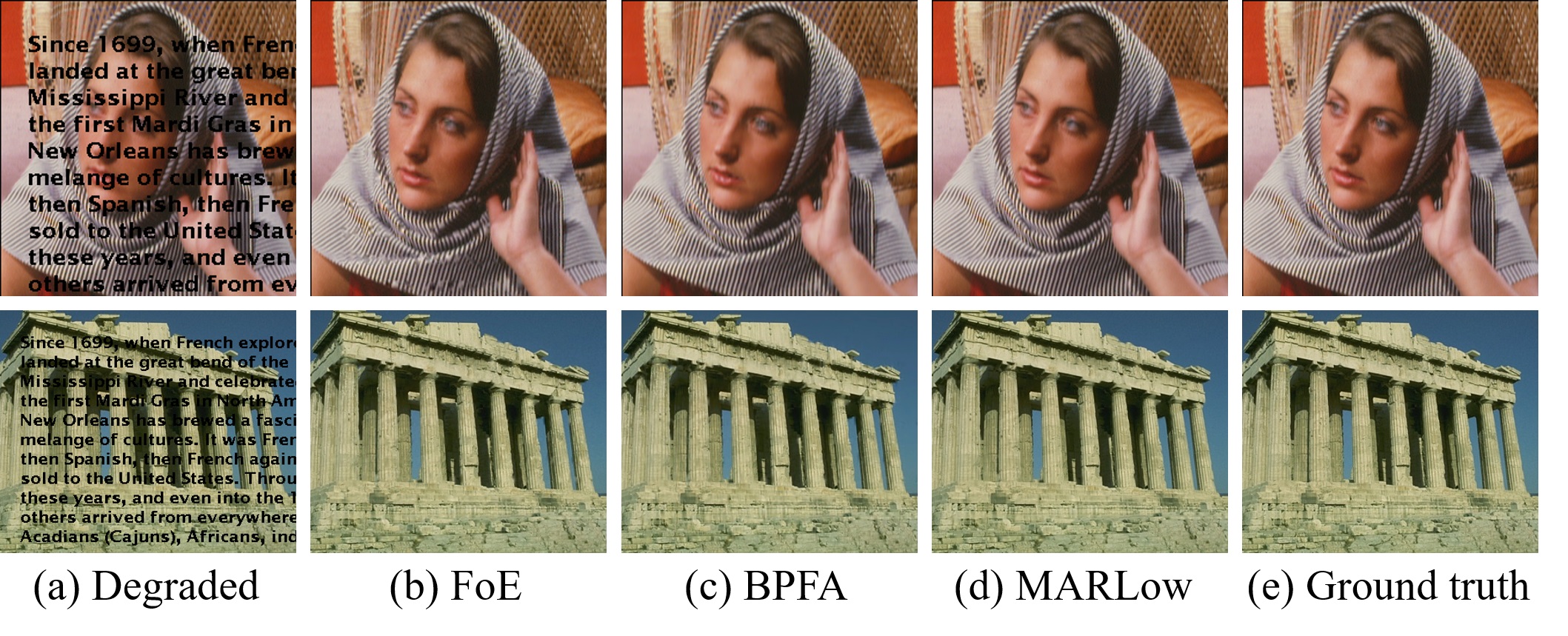}
\vspace{-4mm}
  \caption{Visual quality comparison of text removal for image \emph{Barbara} and \emph{Parthenon}. (a) The degraded image with text mask; (b) - (d) Restoration results by FoE, BPFA, and the proposed method. (e) The ground truth image.}
  \label{text-removal}
\vspace{-6mm}
\end{figure*}

Text removal is one of the classic case of image restoration. The purpose of text removal is to recover the original image from a degraded version by removing the text mask. We have compared our method with four state-of-the-art algorithms: KR \cite{takeda2006robust}, FoE \cite{roth2009fields}, JSM \cite{zhang2014image} and BPFA \cite{zhou2012nonparametric}. Our experimental settings of text removal are the same with those in color image restoration. Tab. \ref{textTable} shows the PSNR and SSIM results of different methods. Fig. \ref{text-removal} presents visual comparison of different approaches, which further illustrates the effectiveness of our method.

\begin{table}[htbp]
  \centering
  \caption{PSNR (dB) and SSIM results of text removal from different methods. The best result in each case is highlighted in bold.}
    \begin{tabular}{|c|c|c|c|c|c|}
    \hline
    \textbf{Image}      & \textbf{KR} & \textbf{FoE} & \textbf{BPFA} & \textbf{JSM} & \textbf{Proposed} \\
    \hline
    \hline
    \textit{Barbara}    & 29.59/0.9578 & 30.18/0.9585 & 32.91/0.9647 & 36.56/0.9839 & \textbf{37.81}/\textbf{0.9862} \\
    \hline
    \textit{Parthenon}  & 29.69/0.9374 & 31.87/0.9535 & 31.90/0.9506 & 33.07/0.9631 & \textbf{33.27}/\textbf{0.9656} \\
    \hline
    \textit{Butterfly}  & 30.22/0.9717 & 30.04/0.9713 & 30.07/0.9595 & 31.85/0.9797 & \textbf{33.18}/\textbf{0.9844} \\
    \hline
    \textit{Foreman}    & 40.51/0.9848 & 38.81/0.9863 & 38.53/0.9733 & 39.70/0.9870 & \textbf{43.30}/\textbf{0.9887} \\
    \hline
    \hline
    \textbf{Average}    & 32.50/0.9629 & 32.73/0.9674 & 33.35/0.9620 & 35.30/0.9784 & \textbf{36.89}/\textbf{0.9812} \\
    \hline
    \end{tabular}%
\vspace{-6mm}
  \label{textTable}%
\end{table}%

\subsection{Image Interpolation}

\begin{figure*}[hbtp]
\vspace{-3mm}
  \centering
  \includegraphics[width=120mm]{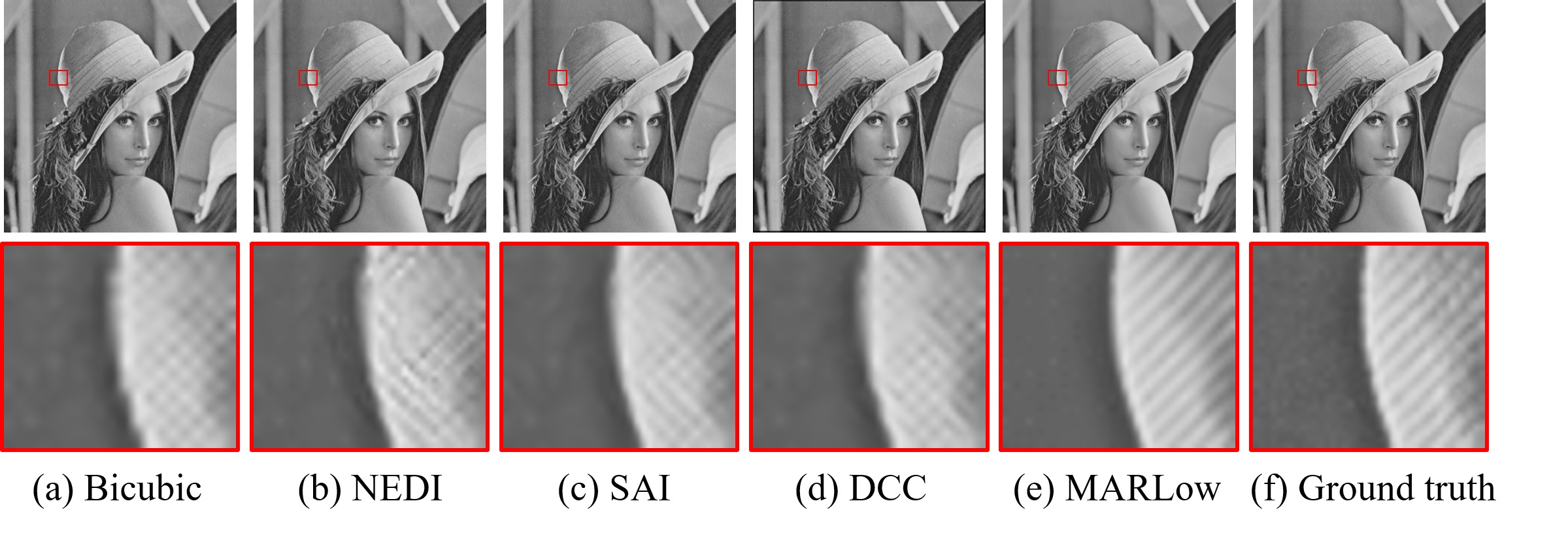}
\vspace{-4mm}
  \caption{Subjective comparison of interpolation for image \emph{Lena}. Results of (a) Bicubic, (b) NEDI, (c) SAI, (d) DCC, (e) the proposed method, and (e) the ground truth image.}
  \label{interpolation}
\vspace{-6mm}
\end{figure*}

The proposed method can also be applied on basic image processing problems, such as image interpolation. In fact, image interpolation can be regarded as a special circumstance of image restoration from random samples. To be more specific, locations of the known/missing pixels in image interpolation are fixed. Since our method is designed to deal with image restoration from random samples, we do not utilize this feature in our current implementation. Even so, we evaluate the performance of the proposed method with respect to image interpolation by comparing with other state-of-the-art interpolation methods. The compared methods including AR model based interpolation algorithms NEDI \cite{li2001new} and SAI \cite{zhang2008image}, and a directional cubic convolution interpolation DCC \cite{zhou2012image}. Objective results are given in Tab. \ref{tab:interpolation} and subjective comparisons are demonstrated in Fig. \ref{interpolation}, showing that proposed method is competitive with other methods.

\begin{table}[htbp]
  \centering
  \caption{PSNR (dB) and SSIM results of interpolation from different methods. The best result in each case is highlighted in bold.}
    \begin{tabular}{|c|c|c|c|c|c|}
    \hline
    \textbf{Images} & \textbf{BC} & \textbf{NEDI} & \textbf{SAI} & \textbf{DCC} & \textbf{Proposed} \\
    \hline
    \hline
    \textit{Cameraman} & 25.37/0.8629 & 25.42/0.8626 & 25.88/0.8709 & \textbf{25.92}/\textbf{0.8731} & 25.36/0.8664 \\
    \hline
    \textit{Lena} & 33.94/0.9149 & 33.78/0.9142 & 34.71/0.9193 & 34.50/0.9197 & \textbf{34.90}/\textbf{0.9225} \\
    \hline
    \textit{Lighthouse} & 26.93/0.8436 & 26.37/0.8386 & 26.65/0.8445 & 27.19/\textbf{0.8483} & \textbf{27.22}/0.8462 \\
    \hline
    \textit{Monarch} & 31.86/0.9561 & 31.78/0.9555 & 33.02/0.9623 & 32.92/0.9623 & \textbf{33.16}/\textbf{0.9640} \\
    \hline
    \hline
    \textbf{Average} & 29.52/0.8944 & 29.33/0.8927 & 30.07/0.8992 & 30.13/\textbf{0.9009} & \textbf{30.16}/0.8998 \\
    \hline
    \end{tabular}%
\vspace{-5mm}
  \label{tab:interpolation}%
\end{table}%

\section{Conclusion}
\label{conclusion}
In this work, we introduce the new concept of the multiplanar model, which exploits the cross-dimensional correlation in similar patches collected in a single image. Moreover, a joint multiplanar autoregressive and low-rank approach for image completion from random sampling is presented, along with an alternating optimization algorithm. Our image completion method can be extended to multichannel images by utilizing the correlation in different channels. Extensive experiments on different applications have demonstrated the effectiveness of our method. Future works include the extensions on more other applications, such as video completion and hyperspectral imaging. We are also interested in adaptively choosing the size of the processing image patch since it might improve the completion result.

\bibliographystyle{splncs}
\bibliography{template-new}
\end{document}